\documentclass[runningheads]{llncs}
\usepackage{graphicx}
\usepackage{amsmath,amssymb} 
\usepackage{color}

\usepackage{graphicx}
\usepackage{amssymb}
\usepackage{bbm}
\usepackage{figsize}
\usepackage{algorithm}
\usepackage{algpseudocode}
\usepackage{algorithmicx}
\usepackage{braket}
\usepackage{multirow}
\usepackage{pifont}
\newcommand{\cmark}{\ding{51}}%
\newcommand{\xmark}{\ding{55}}%

\newlength\mylen

\algnewcommand\algorithmicinput{\textbf{INPUT:}}
\algnewcommand\INPUT{\item[\algorithmicinput]}

\usepackage[breaklinks=true,bookmarks=false]{hyperref}
\begin{document}

\title{Detecting Text in the Wild with Deep Character Embedding Network} 
\titlerunning{Detecting Text with CENet} 


\author{Jiaming Li$\rm u^\star$ \and
Chengquan Zhang\thanks{these authors contribute equally in this work.} \and
Yipeng Sun \and
Junyu Han \and
Errui Ding}
%

\authorrunning{J. Liu, C. Zhang et al.} 


\institute{Baidu Inc, Beijing, China.\\
\email{\{liujiaming03,zhangchengquan,yipengsun,hanjunyu,dingerrui\}@baidu.com}}

\maketitle

\begin{abstract}
Most text detection methods hypothesize texts are horizontal or multi-oriented and thus define quadrangles as the basic detection unit. However, text in the wild is usually perspectively distorted or curved, which can not be easily tackled by existing approaches. In this paper, we propose a deep character embedding network (CENet) which simultaneously predicts the bounding boxes of characters and their embedding vectors, thus making text detection a simple clustering task in the character embedding space. 
The proposed method does not require strong assumptions of forming a straight line on general text detection, which provides flexibility on arbitrarily curved or perspectively distorted text. For character detection task, a dense prediction subnetwork is designed to obtain the confidence score and bounding boxes of characters. For character embedding task, a subnet is trained with contrastive loss to project detected characters into embedding space. The two tasks share a backbone CNN from which the multi-scale feature maps are extracted. The final text regions can be easily achieved by a thresholding process on character confidence and embedding distance of character pairs. We evaluated our method on ICDAR13, ICDAR15, MSRA-TD500, and Total Text. The proposed method achieves state-of-the-art or comparable performance on all of the datasets, and shows a substantial improvement in the irregular-text datasets, i.e. Total-Text.
\end{abstract}
\section{Introduction}

Optical Character Recognition (OCR) is a long-standing problem that attracts the interest of many researchers with its recent focus on scene text. 
It enables computers to extract text from images, which facilitates various applications, such as scene text translation, scanned document reading, etc. 

As the first step of OCR, the flexibility and robustness of text detection significantly affect the overall performance of OCR system. The goal for text detection algorithms is to generate bounding boundaries of text units as tight as possible.



When dealing with different kinds of text, different text unit should be defined in advance. When detecting text in Latin, the text unit is usually ``word"; while if in Asian language, it is ``text line" instead. Words or lines have a strong prior by their nature. The characters in them tend to usually cluster as straight lines. Therefore, it is natural to define rectangles or quadrangles that wrap text as the objective of detection. This prior has been widely used in many text detection works and achieved promising results\cite{zhou2017east,tian15textflow,Hu17WordSup,tian2016ctpn,ma2017arbitrary,liao2017textboxes,li2017towards,epshtein2010swt,neumann12mser}. 

\begin{figure}
\centering
\SetFigLayout{4}{1}
  \subfigure[]
  {\includegraphics[width=0.2\textwidth, height=0.1\textwidth]{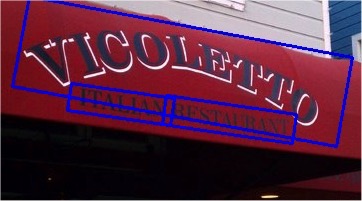}}
  \subfigure[]
  {\includegraphics[width=0.2\textwidth, height=0.1\textwidth]{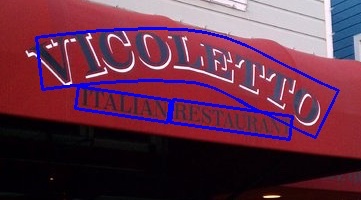}}
   \subfigure[]
   {\includegraphics[width=0.2\textwidth, height=0.1\textwidth]{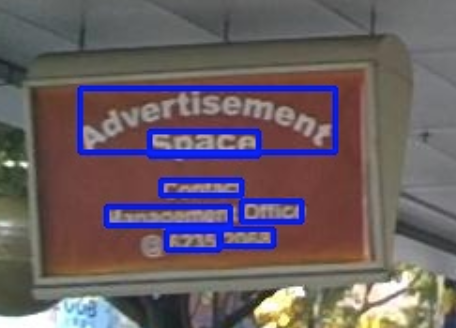}}
  \subfigure[]
  {\includegraphics[width=0.2\textwidth, height=0.1\textwidth]{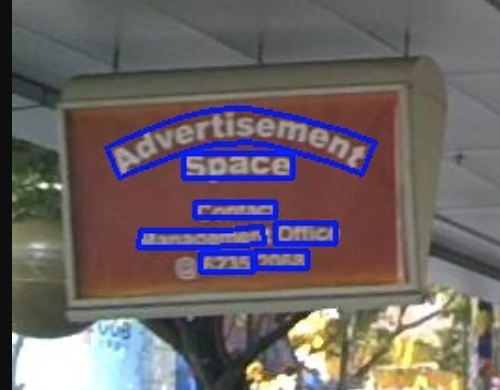}}
\caption{Examples of curved text from Total-Text and ICDAR15. Detected texts are labeled with quadrangle bounding box in (a) and (c), and with polygon by our proposed method in (b) and (d). Note that the image is cropped fobr better presentation. The dark blue lines represent the detected text boundaries.}
\label{fig:curved_text}
\vspace{-1.5em}
\end{figure}

However, 
when text appears in the wild, it often suffers from severe deformation and distortion. Even worse, some text are curved as designed. In such scenario, this strong prior does not hold. Fig. \ref{fig:curved_text} shows curved text with quadrangle bounding boxes and curved tight bounding boundaries. It can be easily observed the quadrangle bounding box inevitably contains extra background, making it more ambiguous than curved polygon boundaries. 


We realized that 
if characters can be detected and a flexible way to group them into text can be found, tight bounding boundaries will be easily generated with the boundary of characters. Characters are also fundamental elements of text, this idea can be naturally extended to irregular text. In early attempts \cite{tian15textflow,yin2009,yin2015multi}, scholars turned to use a heuristic clustering method with hand-crafted features to link detected character parts into text lines. 
The non data-driven heuristic clustering methods are fragile, requiring a thorough check on corner cases manually. Also, the hand-crafted features ignore large parts of visual context information of text, making it less discriminative to determine the closeness between characters. 

Thereby, we propose a Character Embedding Network (CENet) in a fully data-driven way. 
The model detects characters as the first step. After characters being detected, they are projected into an embedding space by the same model where characters belonging to the same text unit are close to each other, and characters belonging to different text units are far from each other. During the training stage, the network is jointly trained with a character detection loss and a character embedding loss. During the inference stage, a single forward pass could produce character candidates as well as their representation in the embedding space. A simple distance thresholding is then applied to determine connected character pairs. Connected character pairs further form text groups by chaining the characters together. After the connection relationships are properly learned, the text units could be detected regardless of text length or distortion the text suffers.

To the best of our knowledge, the proposed CENet is the first to model text grouping problem as a character embedding learning problem. 
It does not rely on strong priors, making it capable of detecting arbitrarily curved or distorted text. Moreover, since both character detection and character grouping tasks are based on local patch of images, our model could be directly expand from ``word" detection to ``line'' detection without modifying the backbone model for larger receptive field. Our model also avoids complicated heuristic grouping rules or hand-crafted features. At last, our single model performs two tasks with a single forward pass, only adding minimal overhead over character detection network. 

The contributions of this paper are three-fold:
\begin{itemize}
\item We propose a multi-task network to detect arbitrarily curved text in the wild. The character detection subnet is trained to detect character proposals, and the character embedding subnet learns a way to project characters into embedding space. Complicated post-processing steps, e.g. character grouping and word partition, are then be simplified as a simple distance thresholding step in the embedding space. 

\item We adopt a weakly supervised method to train character detector with only word-level polygon annotations, without the strong hypothesis that text should appear in a straight line. 

\item We conduct extensive experiments on several benchmarks to detect horizontal words, multi-oriented words, multi-oriented lines and curved words, demonstrating the superior performance of of our method over the existing methods. 

\end{itemize}


\section{Related works}
\textbf{Scene Text Detection} 
Based on the basic elements produced in the pipeline of text detection, we roughly classify the scene text detection methods into three categories: 

\emph{Text Component Based Methods.} MSER~\cite{neumann12mser} and SWT~\cite{epshtein2010swt} are classical text component extraction methods. 
In the era of deep learning, CTPN~\cite{tian2016ctpn} extracts horizontal text components with fixed-size width using a modified Faster R-CNN framework. Horizontal text lines are easily generated, since CTPN adjusted the Faster R-CNN~\cite{ren15faster} framework to output dense text components. SegLink~\cite{shi2017detecting} proposed a kind of oriented text component (i.e. segment) and a component-pair connection structure (i.e. link). A link indicates which two segments should be connected. Naturally, SegLink dealt better with multi-oriented texts than CTPN. PixelLink~\cite{deng2018pixellink} provided an instance segmentation based solution that detects text pixels and their linkage with neighbor pixels. Positive pixels with positive links are grouped as the collection of connected components. Besides, Markov Clustering Network~\cite{liu2018learning} regarded detected text pixels as nodes and associated them with computed attractors by a designed markov clustering networks. 
The above mentioned methods provided inspiring ideas on text detection. However, the regions between characters are sometimes in-discriminative with background in some cases, especially in text lines where distances between characters are large. 

\emph{Character Based Methods.} The character is the fundamental semantic element of text, regardless of whatever the language is. Additionally, characters have a precise definition. Compared with components, the  position, scale and orientation information of characters could be provided in a well-defined way. 
Therefore, 
the character seems to be a more natural choice to set up a general text detection engine. Previously, some character based methods~\cite{zhu2016text,tian15textflow,He16TextAttention,jaderberg2014deep} have achieved good performances on some public benchmarks that have character-level annotations (such as ICDAR13~\cite{karatzas13icdar}). However, as it is not convenient 
and economic to acquire character-level annotations, more and more public benchmarks (such as ICDAR15~\cite{karatzas15icdar} and MSRA-TD500~\cite{yao2012detecting}) provide only word-level annotations. Hence, these methods can not
 get sufficient fine-tuning on those datasets. In order to deal with this problem, 
Tian et al.~\cite{Tian17WeText} (WeText) and Hu et al.~\cite{Hu17WordSup} (WordSup) proposed two different solutions for character detection from word annotations. 

 Unfortunately, both methods can not deal with the datasets (such as Total-Text~\cite{chng17tt}) with polygon annotations for curved text lines, because they are based on the strong hypothesis that text should appear in a straight line.  Our method is also a character-based method with weak supervision, but a more general mechanism of updating character supervision is proposed, which makes our method capable of learning from arbitrary annotation formats including rectangle, quadrangle and polygon. 

\emph{Word Based Methods.} Recently, quite a few works~\cite{gupta2016synthetic,zhong17icassp,liao2017textboxes,zhou2017east,liu2017deep,he2017direct,he2017single,lyu2018multi} have put emphasis on adjusting some popular object detection frameworks including Faster R-CNN~\cite{ren15faster}, SSD~\cite{liu2016ssd} and Densebox~\cite{huang2015densebox} to detect word boundary. In contrast to general objects, texts appearing in the real-world have larger varieties of aspect ratios and orientations. Liao et al.~\cite{liao2017textboxes} and Zhong et al.~\cite{zhong17icassp} directly added more anchor boxes of large aspect ratio to cover texts of wider range. Gupta et al.~\cite{gupta2016synthetic} and Liu et al.~\cite{liu2017deep} added the angle property to the bounding box to deal with the problem of multiple orientations, while EAST~\cite{zhou2017east} and He et al.~\cite{he2017direct} provided a looser representation namely quadrangle. These methods seem to easily achieve high performance on benchmarks with word-level annotations, but not on non-Latin scripts or curved text with polygon-level annotations.


\textbf{Deep Metric Learning.} The goal of metric learning or embedding methods~\cite{chopra2005contrative,schroff2015facenet,Wang17angular} is to learn a function that measures how similar two samples are. There are many successful applications of metric learning~\cite{hoi2006learning,chopra2005contrative,schroff2015facenet,Wang17angular}, such as ranking, image retrieval, face verification, speaker verification and so on.  
By far, applications of metric learning on document analysis or text reading were limited to the problem of word spotting and verification~\cite{almazan2014word,ren15faster,tomas17ctrl}. 
In this work, we verify the effectiveness of deep metric learning in text detection task. Based on character candidates, we provide an end-to-end trainable network that can output the character bounding boxes and their embedding vectors simultaneously. Text regions could be easily detected by grouping characters which embedding distances are small.

\section{Method}
There are two tasks that our model is supposed to solve. One is to detect characters and the other is to project characters into an embedding space where characters belonging to the same group are close, and characters belonging to different groups are far from each other. Sharing a backbone CNN, the two tasks are implemented by separate subnets, i.e., a character detection subnet and a character embedding subnet. To put it another way, our framework is a single backbone network with two output heads. 
With the calculated character candidates and their corresponding embedding vectors, the post processing removes false positive and groups characters in an efficient and effective manner.

\subsection{Network design}
\label{sec:net_design}
We use ResNet-50\cite{he2016deep} as the backbone network of our model. Following recent network design practices \cite{tian15textflow,fpn,Hu17WordSup}, we concatenate semantic features from three different layers of the backbone ResNet-50 network. After deconvolutional operations, the features are concatenated as shared feature maps which are 1/4 of the original image in size. A character detection subnet and a character embedding subnet are stacked on top of the shared feature maps.

The character detection subnet is a convolutional network that produces 5 channels as the final output. The channels are offsets $\Delta x_{tl}$, $\Delta y_{tl}$, $\Delta x_{br}$, $\Delta y_{br}$ and confidence score, where $tl$ means top left and $br$ means bottom right. The top left and bottom right bounding box coordinates of detected character candidates could be calculated by $(x-\Delta x_{tl},\, y-\Delta y_{tl})$ and $(x+\Delta x_{br},\, y+\Delta y_{br})$, where $x$ and $y$ are coordinates of pixel whose confidence score greater than a threshold $s$. The bounding boxes further serve as RoIs of characters. 

The character embedding subnet takes the residual convolution unit (RCU) as the basic blocks which is simplified residual block without batch normalization. 
The design was inspired by \cite{tian15textflow} where the authors showed that the scores and bounding box sizes of character proposals offer strong clues on whether they belong to the same group, and the feature maps extracted by the backbone network contains such information. Therefore, residual units were chosen to preserve score and bounding box information from feature maps, directly passing them to top layers by skip connection. On top of the  RCU blocks, we employ a 1x1 convolution layer with linear activation function to output a 128-channel final embedding map. RoI pooing with $1\times 1$ kernel is applied on the embedding maps extracting embedding vectors for each character.

\begin{figure*}
\begin{center}
\includegraphics[width=\textwidth]{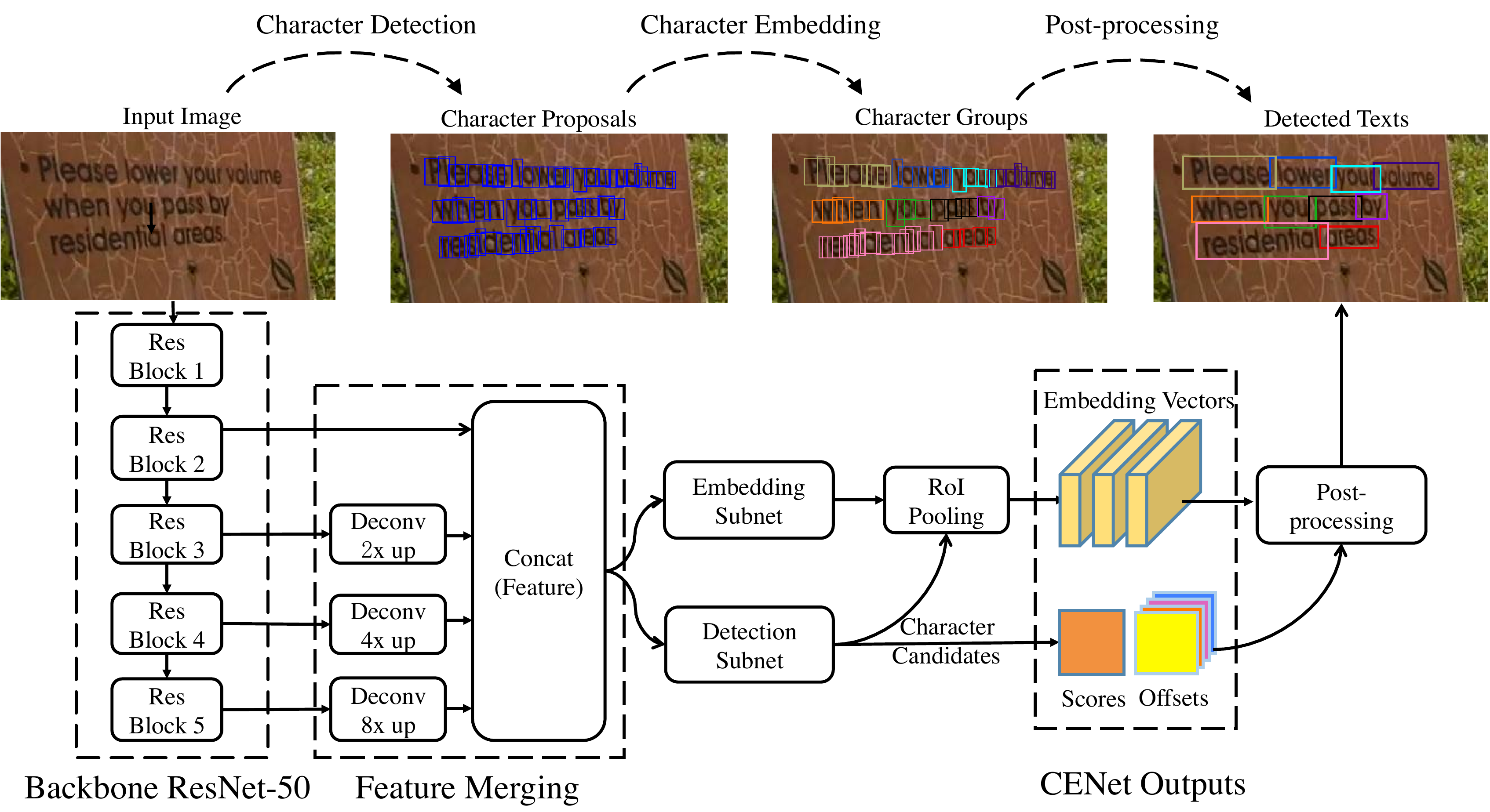}
\caption{Overall process of the model. Blue bounding boxes in ``character proposals" are character candidates with high confidence scores. ``Character Clusters" is the character clusters in embedding space, where candidates in the same cluster use the same color. The final detected words represented in quadrangles are shown in ``Detected text". Better view in color.}
\label{fig:overall}
\end{center}
\vspace{-1.5em}
\end{figure*}

During inference, we extract confidence map, offset maps and embedding maps from the two heads of the model. After thresholding on the score map and performing NMS on character proposals, the embedding vectors are extracted by $1\times 1$ RoI pooling on embedding map. In the end, we output character candidates with the format of $\{$score, coordinates($x, y$) of character center, width, height, 128D embedding vector$\}$. Characters are finally clustered into text blocks as the last post-processing step. The overall structure of the model and pipeline are shown in Fig. \ref{fig:overall}.

\subsection{Training character detector}
\subsubsection{Loss definition}
The character detector consists of two tasks that include text/non-text classification and box regression. The loss can be formulated as
\begin{equation}
\label{equ:char_loss}
\mathit{L_{char}}= L_{cls} + {\lambda}_{1} L_{reg}, 
\end{equation}
where $L_{cls}$ denotes the binary classification loss, $L_{reg}$ represents the box regression loss, and ${\lambda}_{1}$ is a factor to balance the two losses. In this paper, we use pixel-wise hinge-loss as classification cost. Some measures for class balance or boosting (e.g., OHEM~\cite{shrivastava2016ohem}) are adopted in our experiments. Usually, we set the sampling ratio of $1:3$ to balance the positive and negative samples, where $30\%$ of negative samples selected from the top hardest in a training batch. Here, IoU-loss~\cite{yu2016unitbox} is adopted as the regression cost which handles the problem of bounding box accuracy bias between large and small objects instead of L2-loss. 

\subsubsection{Learning character detector from coarse annotation}
Since it is labor-intensive to annotate character-level boxes, most of public benchmarks like ICDAR15~\cite{karatzas15icdar} and Total-Text~\cite{chng17tt} provide only quadrangle or polygon annotations for words, and MSRA-TD500 provides annotations for sentences. Those annotations are all coarse annotations. 
Inspired by WordSup~\cite{Hu17WordSup}, which recursively rectifies character-level supervisions and updates the model parameters with the rectified character-level supervision, a new rectification rule is designed for producing character-level supervision. 
This rule is capable of training character detector from bounding boundary annotations with polygon format, while WordSup may fail. 

\begin{figure*}
\centering
\SetFigLayout{3}{1}
  \subfigure[]  {\includegraphics[width=0.24\textwidth]{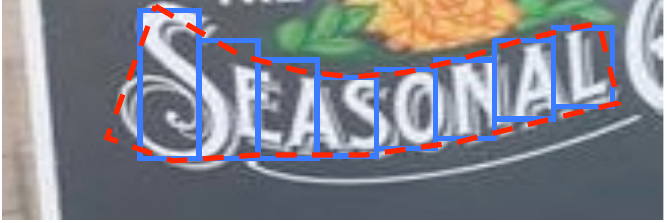}}
  \subfigure[]  {\includegraphics[width=0.24\textwidth]{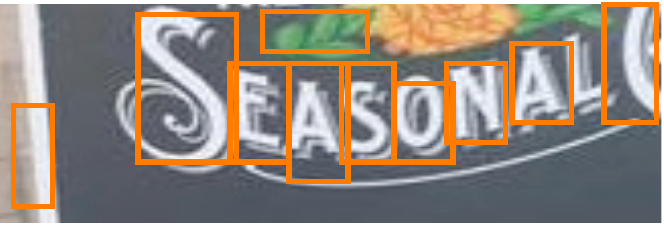}}
  \subfigure[]  {\includegraphics[width=0.24\textwidth]{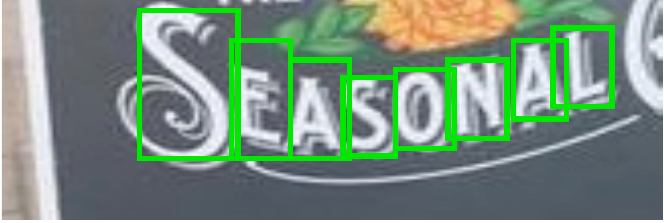}}
\caption{Learning character detector from word-level annotation. (a) are some coarse-char boxes (blue) with the polygon annotation (red), (b) are some pred-char boxes, and (c) are the fine-char boxes whose height is the same as (a).}
\label{fig:week_char}
\vspace{-1.5em}
\end{figure*}


Our design follows the general observation that the short side of a nearly horizontal (or vertical) text is approximately equal to the heights (or width) of characters in it. The short side could be used to rectify the imprecise predicted characters with the following pipeline. 
Firstly, each annotated quadrangle or polygon is uniformly divided into $N$ bounding boxes along the center line, where $N$ denotes the character number of the text transcript. We call the preliminary bounding box segmentations as coarse-char boxes. After one forward pass, some candidate character boxes (namely pred-char boxes) with high confidence are collected.  Finer character boxes (namely fine-char boxes) are produced from coarse-char boxes and their corresponding matched pred-char boxes. If no matched pred-char is founded, the coarse-char box is used as a fine-char box directly. Otherwise, if the annotated text is more horizontal, the width of the fine-char box is set to be the width of pred-char box, and the height is set to the height of the coarse-char box; if more vertical, the width is the width of coarse-char box, and the height is the height of pred-char box. The obtained fine-char boxes are used as ``ground truth" in Equ. \ref{equ:char_loss} to update model.

The matched pred-char box $p$ of a coarse-char box $c$ should meet the following constraints:
\begin{equation}
\label{equ:char_up}
\left\{
\begin{aligned}
S(p) > t_{1} \\
IoU(p, c) > t_{2}, \\
\end{aligned}
\right.
\end{equation}
where $S(p)$ denotes the confidence score of pred-char box $p$, $IoU(p, c)$ means Intersection over Union between the pred-char box and coarse-char box. $t_{1}$ and $t_{2}$ are predefined to 0.2 and 0.5 in our experiments. The visualization of the rectification procedure is shown in Fig.~\ref{fig:week_char}.

  
    

    



\subsection{Learning character embedding}
The character embedding subnet is another crucial part in our model. 
In an ideal case, we hope the subnet projects the characters into an embedding space. Distances between characters among the same text unit are small in the learned space, and that between those belong to different units to be large. Therefore we can group characters into text blocks by performing clustering in the embedding space. This case resembles the objective of metric learning, which aims to learn a distance function to measure similarities between samples. 

Inspired by previous works in metric learning, we select the most straightforward contrastive loss to train our model. 
Contrastive loss takes $pairs$ of characters into calculation. Let $i$ and $j$ denote the index of character candidates in a pair, $v_i$ and $v_j$ denote their embedding vectors that are extracted by the embedding subnet, and $l_{i,j}$ denotes whether they belong to the same text unit. If they do, we name pair ${(i,j)}$ to be positive pair and $l_{i,j} = 1$. Otherwise, pair ${(i,j)}$ is defined as negative pair and $l_{i,j} = 0$ . The Contrastive Loss is defined as 
\begin{equation}
\label{equ:contrastive}
J(i, j)= l_{i,j}[D(v_i,v_j)]^2 + (1-l_{i,j})\textrm{max}(0,1 - D(v_i,v_j))^2 ,
\end{equation}
where $D$ denotes the distance measure. In training, $v_i$ and $v_j$ are pulled close to each other if $l_{i,j}=1$. If $l_{i,j}=0$, $v_j$ and $v_i$ are pushed away from each other until $D(v_i,v_j)>1$. 


\subsubsection{Constructing Local Character Pairs}
\label{sec:reweight}

It is worth-noting that in every definition of text, characters in the same text unit are naturally close in the image. Two small characters are unlikely from the same text unit if they are too far from each other in the image. However, if they are on the endpoints of a line of characters, the probability of their belonging to same text line are significantly increased. The key difference is whether there are closely scattered characters, namely \textit{local character pairs}, that connect individual characters in one text unit.

In addition, it is unnecessary to train models with all possible character pairs. Instead, when all the local character pairs are correctly labeled, all of the text units would be correctly detected. Working with local character pairs also reduces the requirement of large receptive field when detecting long text. 

In this work, we employ k nearest neighbor with radius (\textrm{r-KNN}) to incorporate such information. When producing possible pairs, each character was selected as $anchor$ in turn. 
With an $anchor$ selected, at most $k$ characters which are closest to $anchor$ in the image were taken form $pairs$. 
Another useful heuristic rule is that a character is more likely to be connected with characters with similar box size. Therefore, only characters within radius were kept. To formalize this empirical pair sampling method, we define $c_i$, $w_i$, and $h_i$ as the center coordinates, width, and height of character $i$ in image respectively; and \textrm{KNN}(i) be a function that generates the set of k nearest neighbors of character $i$ in the image. Then $j \in \textrm{r-KNN}(i,\beta r(i))$ represents $j$ is in \textrm{KNN}(i) and the spatial distance $D(c_i,c_j)<\beta \sqrt{w_i^2+h_i^2}$. Both $k$ and $\beta$ were set to 5 in our experiments.

When $j \in \textrm{r-KNN}(i)$, we call $i$ and $j$ produces a locally connected pair. Here we define the set of all locally connected pairs as $LCP = \{(i,j), i \in M, j \in \textrm{r-KNN}(i)\}$, where M is the total number of character candidates in one image. With \textrm{r-KNN} preprocessing, there are only $O(kM)$ locally connected pairs remaining, reducing the size of character pairs to a reasonable level. 



We noticed that the positive pairs are redundant. The minimum requisite for error-less positive pairs is that at least one chain connects all characters in a text unit. Positive pairs with large embedding distances do not contribute any text level error as long as the minimum requisite is satisfied. However, a negative pair with small embedding distance will certainly mis-connect two text units and generate text level error. Meanwhile, we found there are about 3/4 of local character pairs are positive. According to the above analysis, we assume the negative pairs should be weighted more than the positive pairs in training. 

Therefore, we sample $R$ pairs from $LCP$ of batch images so that there are $\alpha$ pairs are negative in a batch. Let's denote the sampled pairs set as $SP$, the final re-weighted loss for learning embedding is defined as Equ. \ref{equ:final_loss}. We found $R=1024$ and $\alpha=60\%$ work well in our experiments. 

\begin{equation}
\label{equ:final_loss}
\mathit{L_{emb}}= \frac{ \sum_{i,j \in SP}J(i,j)}{R}.
\end{equation}

The loss function to train the whole network then becomes

\begin{equation}
\label{equ:overall_loss}
\mathit{L}= \mathit{L_{cls}} + \lambda_1 \mathit{L_{reg}} + \lambda_2 \mathit{L_{emb}},
\end{equation}
where $\lambda_1$ and $\lambda_2$ control the balance among the losses. We set both $\lambda_1$ and $\lambda_2$ to 1 in our experiments.

\subsection{Post-processing}
In testing, we employ two threshold values ($s$ and $d$) to filter false character candidates and group characters into text units. After a forward pass, the proposed model would provide a set of character candidates and their corresponding embedding vectors. Then, the character candidates with confidence scores greater than $s$ are kept. Next, \textrm{r-KNN} is performed on each character, outputting the local character pairs in whole image. To address the character grouping problem, we simply cut down the connected pairs whose embedding distances are over $d$. 

Following the steps above, we can quickly find characters from the same groups. The final step is to represent the character groups in a suitable way. In this paper, we adopted the piecewise linear method that used in WordSup~\cite{Hu17WordSup} to format the boundary of character groups. This method provides various configurable boundary formats, which meet the requirements of different benchmarks. On ICDAR15, a filtering strategy that removes short words with less than two detected characters are applied. This strategy aims to further remove false alarm from the detection results.



\section{Experiments}

We conduct experiments on ICDAR13, ICDAR15, MSRA-TD500, and Total-Text datasets, to explore how the proposed approach performs in different scenarios. The four chosen datasets focus on horizontal-oriented text, multi-oriented text, sentence-level long text, as well as curved-oriented text respectively. Experiments on synthetic data are also conducted for 
structural search and 
pretraining. We also list recent state-of-art methods for comparison.

\subsection{Datasets and Evaluation}
Five datasets are used in the experiments:
\begin{itemize}
\item VGG 50k. The VGG SynthText dataset \cite{gupta2016synthetic} consists of 800,000 images, where the synthesized text are rendered in various background images. The dataset provides detailed character-level, word-level and line-level annotations. For the experimental efficiency, we randomly select 50,000 images for training and 500 images for validation. This subset is referred as VGG 50k.

\item ICDAR13. The ICDAR13 dataset \cite{karatzas13icdar} is from ICDAR 2013 Robust Reading Competition. The texts are well focused and horizontal oriented. Annotations on character-level and word-level bounding boxes are both provided. There are 229 training images and 233 testing images. 

\item ICDAR15. The ICDAR15 dataset \cite{karatzas15icdar} is from ICDAR 2015 Robust Reading Competition. The images are captured in an incidental way with Google Glass. Only word-level quadrangles annotations are provided in ICDAR15. There are 1000 natural images for training and 500 for testing. 
Experiments under this dataset shows our method's performance in word-level Latin text detection task. 

\item MSRA-TD500. The MSRA-TD500 dataset \cite{yao2012detecting} is a dataset comprises of 300 training images and 200 test images. Text regions are arbitrarily orientated and annotated at sentence level. Different from the other datasets, it contains both English and Chinese text. We test our method on this dataset to show it is scalability across different languages and different detection level (line level in this dataset). 

\item Total-Text. The Total-Text dataset \cite{chng17tt} is recently released in ICDAR2017. Unlike the ICDAR datasets, there are plenty of curved-oriented text as well as horizontal and multi-oriented text in Total-Text. There are 1255 images in training set, and 300 images in test set. Two kinds of annotations are provided: one is word level polygon bounding regions that bind ground-truth words tightly, and word level rectangular bounding boxes as other datasets provided. 
 Since many of the words in this datasets are curved or distorted, it is adopted to validate the generalization ability of our method on irregular text detection tasks.
\end{itemize}

\subsection{Implementation details}

Since the training samples are not abundant in these available datasets, we use VGG 50k data to pretrain a base model, and then finetune the base model on other benchmark datasets accordingly. Two models are trained with the word-level annotation and line-level annotation of VGG 50k data respectively. The backbone ResNet-50 model was first pretrained on ImageNet dataset. Then the models are trained on VGG 50k dataset for character detection and further finetuned with both character detection and character embedding loss. The converged models are used as pretrained models for training other benchmarks. 



We have not adopted any more data augmentation when training models with VGG 50k data. For the remaining benchmark datasets, we perform multi scale data augmentation by resizing the image to [0.65, 0.75, 1, 1.2] scales of the original image respectively, and cropped with a sliding window of size $512 \times 512$ with stride $256$ to generate images for training. 
During training, we randomly rotate the cropped image to $90^o$, $180^o$ or $270^o$, and distort brightness, contrast and color on all three benchmark datasets.

When training with data without character level annotation, the supervision for character detection comes from the weak supervision mechanism depicted above. Boxes used to train character embedding are the same coarse-char box used for character detection. We found a ``mixing batch" trick helps. In practice, a half of the mixing batch are sampled from benchmark data, and the other half are from VGG 50k which provide character-level annotation. Character supervision for data from VGG 50k comes from their character annotation. 

The optimizer is SGD with momentum in all the model training. We train the models 50k iteration at learning rate of 0.001, 30K iterations at 0.0001, and 20K iterations at 0.00001. The momentum was set to 0.9 in all the experiments. 
The two threshold for post-processing, i.e. $s$ and $g$, are tuned by grid search on training set.

All the experiments are carried out on a shared server with a NVIDIA Tesla P40
GPU. Training a batch takes about 2s. Inference was done on original images. The average inference time cost is 276ms per image with size $768 \times 1280$, the forward pass, \textrm{r-KNN} search, NMS, and other operations cost 181ms, 21ms, 51ms and 23ms, respectively.

\subsection{Ablation Study}
As shown in Tab. \ref{tab:ablation}, ablation experiments have been done on ICDAR15 dataset. Three key components in our pipeline are evaluated. Specifically, the mixing batch trick used in weak supervision, the positive-negative pair reweighting strategy, and short word removal strategy are added progressively to show their impact on overall performance.  


\begin{table}
\setlength{\abovecaptionskip}{-10pt}
\setlength{\belowcaptionskip}{10pt}
\begin{center}
\caption{Text detection results using CENet evaluated on the ICDAR15\cite{karatzas15icdar}. }
\begin{tabular}{|c|c|c|c|c|c|}
\hline
  mixing batch trick & reweighting & short word removal &  Recall & Precision & F1 \\
\hline
\xmark & \xmark  &  \xmark & 73.8 & 80.5 & 77.0 \\
\hline
\cmark & \xmark  &  \xmark & 77.4 & 79.4 & 78.4 \\
\hline
\cmark & \cmark  &  \xmark & 79.2 & 82.3 & 80.9 \\
\hline
\cmark & \cmark  &  \cmark & 79.2 & 86.1 & 82.5 \\
\hline
\end{tabular}
\label{tab:ablation}
\end{center}
\end{table}

Without bells and whistles, the model trained merely with weak character supervision and local character pairs converges successfully but gives mediocre results (73\% in Recall). The character detection subnet was more likely overfitted on text components instead of characters. 

With ``mixing batch" trick, word recall is improved strikingly by about 4\% with similar precision. The finding here may imply that this trick, as a regularization measure, prevents the weak character supervision from prevailing. In other words,  weak character supervision tends to results in a certain amount of ``soft" ground truths while the precise character supervision can pull the trained model to its correct position.

If we further add positive-negative pair reweighting trick in character embedding, performances in both precision and recall increase by 2\%. In accordance to our previous analysis in Sec.\ref{sec:reweight}, more balanced positive-negative pairs are behind the improvement.
In addition, a detected word is error-prone if it is too short. Removal of the word less than 2 characters is adopted, which indicates 3.8\% improvement in precision without hurting recall. 

\subsection{Experiments on Scene Text Benchmarks}

\begin{figure*}
\centering
\SetFigLayout{4}{2}

  \subfigure[]  {\includegraphics[width=0.24\textwidth, height=0.15\textwidth]{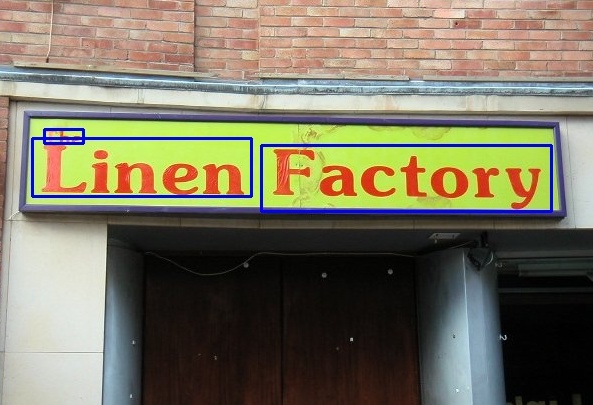}}
  \subfigure[]  {\includegraphics[width=0.24\textwidth, height=0.15\textwidth]{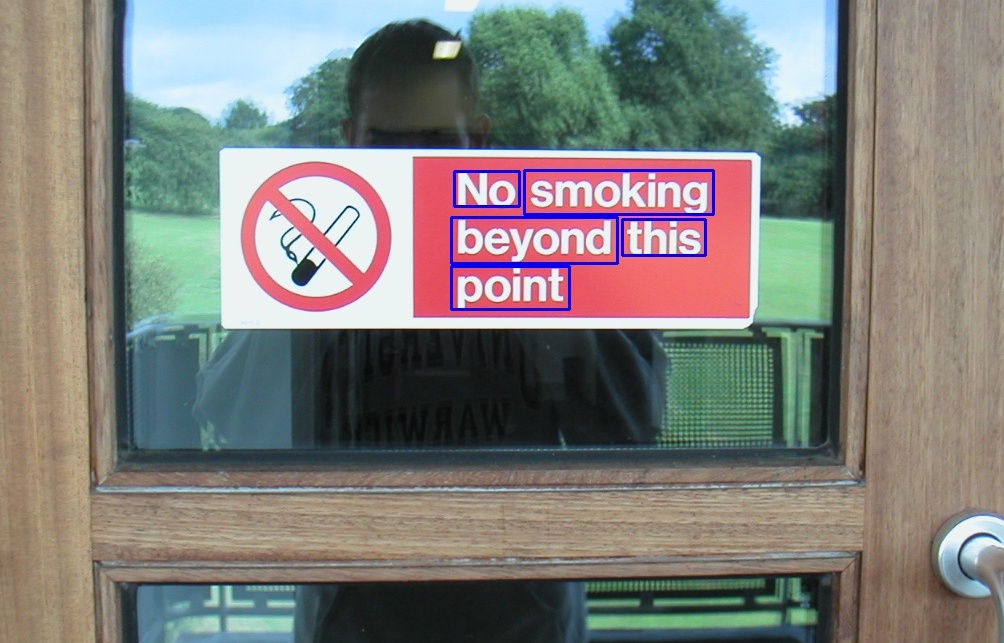}}
    \subfigure[]  {\includegraphics[width=0.24\textwidth, height=0.15\textwidth]{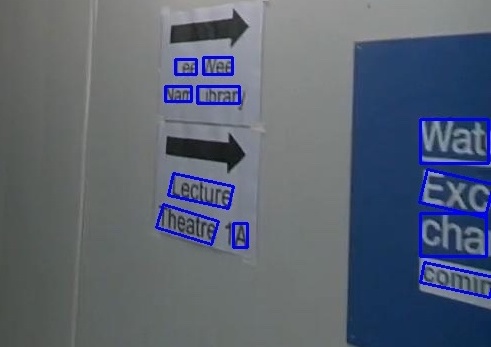}}
  \subfigure[]  {\includegraphics[width=0.24\textwidth, height=0.15\textwidth]{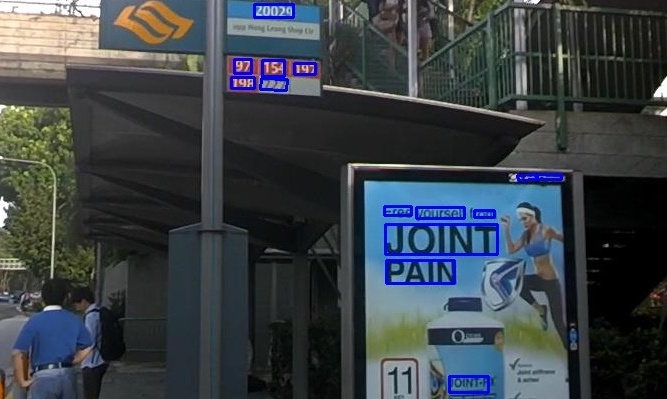}}
  \subfigure[]  {\includegraphics[width=0.24\textwidth, height=0.15\textwidth]{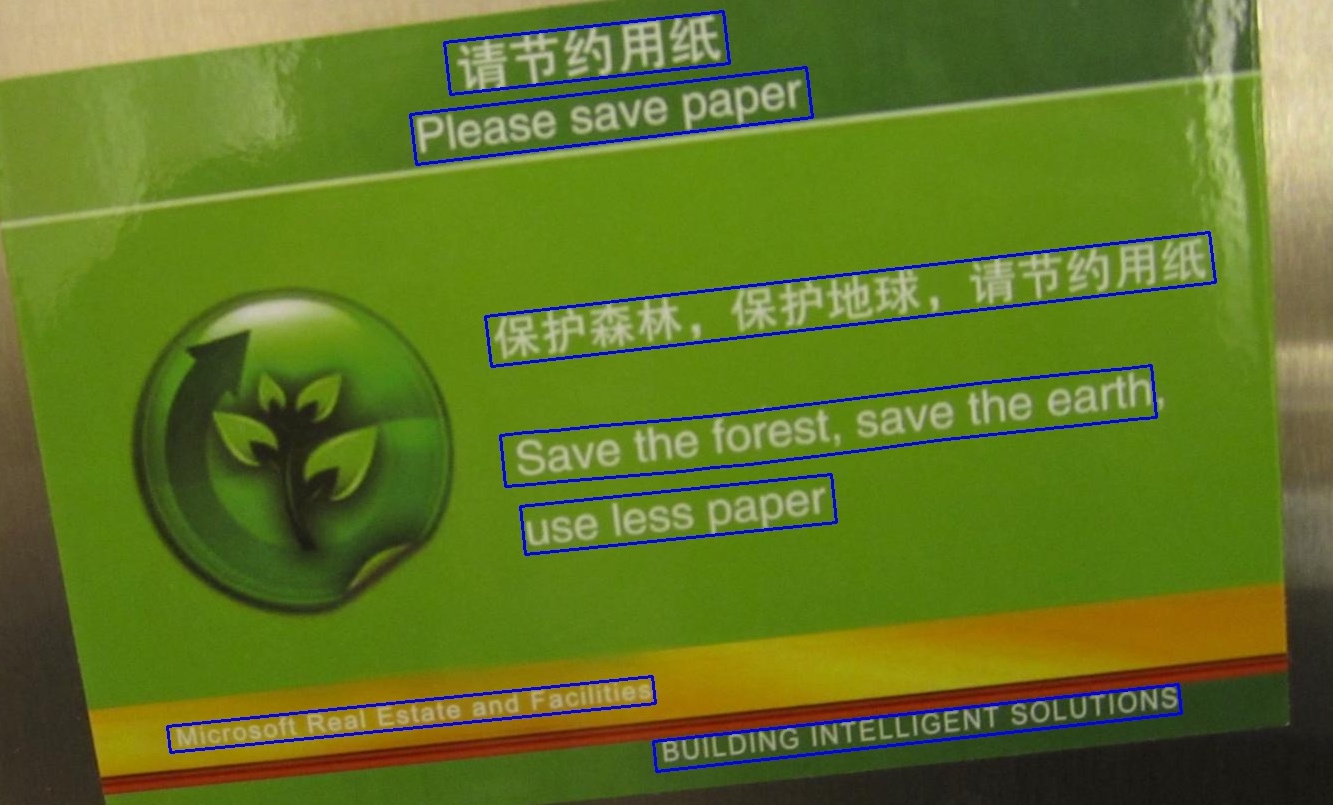}}
  \subfigure[]  {\includegraphics[width=0.24\textwidth, height=0.15\textwidth]{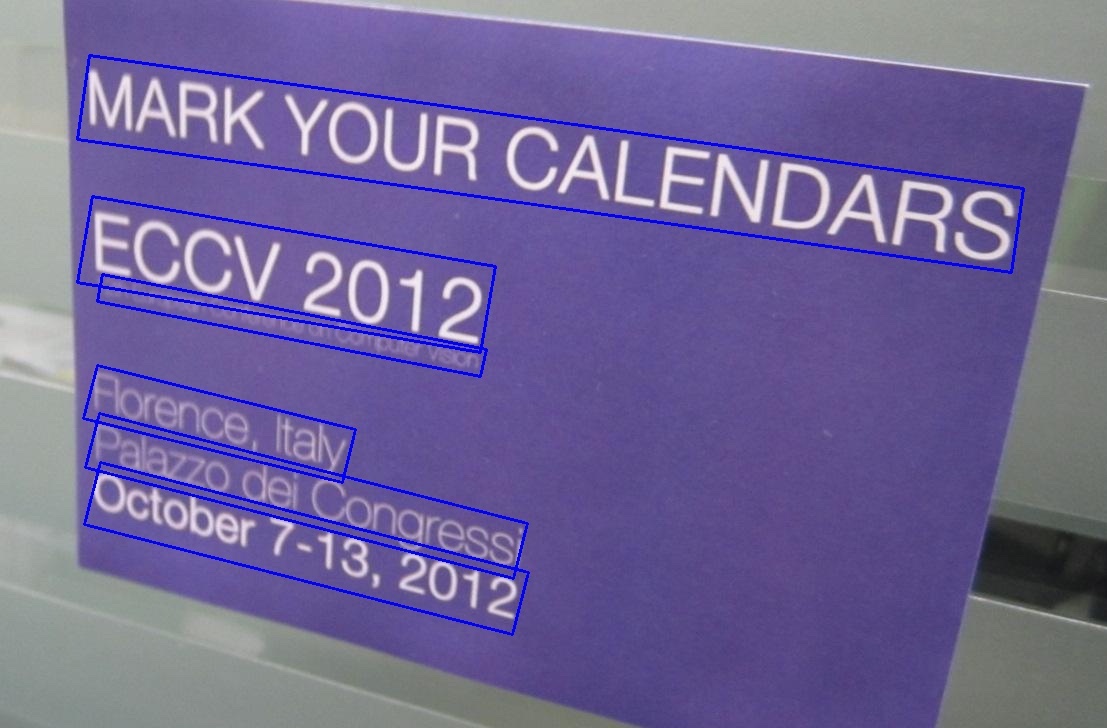}}
  \subfigure[]  {\includegraphics[width=0.24\textwidth, height=0.15\textwidth]{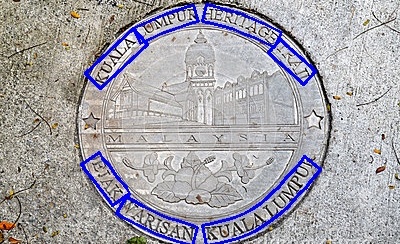}}
  \subfigure[]  {\includegraphics[width=0.24\textwidth, height=0.15\textwidth]{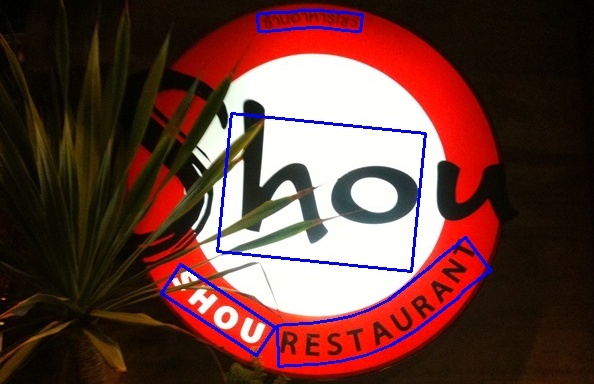}}
  
\caption{Qualitative evaluation of the proposed CENet. Dark blue bounding boundaries show our text detection results on the datasets. Fig. (a,b) are from ICDAR13; fig. (c,d) are from ICDAR15; fig. (e,f) are from MSRA-TD500; and fig. (g,h) are from Total-Text dataset. Zoom in to see better.}
\label{fig:res}
\vspace{-1.5em}
\end{figure*}

Tab.~\ref{tab:13} lists the results on ICDAR13 dataset of various state-of-art methods. Our model presents a competitive performance on this scenario. The demonstrated that the proposed CENet is capable of learning horizontal text line. Note that WordSup adopted the horizontal nature of text directly when grouping characters into text lines, and the data-driven CENet could achieve a similar performance without utilizing that strong prior. $s, d$ are set to 0.4 and 0.45 in this dataset. 

We conduct experiments on ICDAR15 dataset, comparing the results of the proposed CENet with other state-of-the-art methods. As shown in Tab.~\ref{tab:15}, our single scale CENet outperforms most of the existing approaches in terms of F-measure. This shows that character detection and character embedding together can handle most cases in regular text word detection. Our model learns both the character proposals and their relationship in terms of grouping, reducing wrongly-grouped and wrongly-split characters compared with word based methods\cite{zhou2017east,he2017direct}. $s, d$ are set to 0.35 and 0.38 in this dataset.



Tab.~\ref{tab:td500} lists the results on MSRA-TD500 dataset. Our model achieve best result w.r.t F-measure on this dataset. 
The dataset is multilingual and is a good test-bed for generalization. For our model, it is basic unit is character which is dependent on local patch and character embedding connects neighboring units by propagation. Therefore it escapes from the large receptive field requirement of one stage methods. $s, d$ are set to 0.4 and 0.3 in this dataset.



\begin{table}
\begin{center}
\caption{Performances of ICDAR13, the evaluation criteria are from DetEval.}
\begin{tabular}{|c|c|c|c|c|}
  \hline
  Method & Recall & Precision & F-measure \\
  \hline
  \hline
  MCLAB-FCN \cite{zhang2016multi} & 79.65 & 88.40 & 83.80 \\

  Gupta et al. \cite{gupta2016synthetic} & 75.5 & 92.0 & 83.0 \\
  
  Zhu et al. \cite{zhu2016text} & 81.64 & 93.40 & 87.69 \\

  CTPN \cite{tian2016ctpn} & 81.64 & 93.40 & 87.69 \\
  
  WeText \cite{Tian17WeText} & 83.1 & 91.1 & 86.9 \\
  Wordsup~\cite{Hu17WordSup} & 87.53 & 93.34 & \textbf{90.34} \\
   EAST \cite{zhou2017east} & 82.67 & 92.64 & 87.37 \\
 He et al. \cite{he2017direct} &  92 & 81 & 86 \\
 PixelLink \cite{deng2018pixellink} & 88.6 & 87.5   &88.1 \\
  \hline
  CENet (VGG 50k+icdar13 finetune) & 85.94 & 93.18 & 89.41 \\
  \hline
\end{tabular}
\label{tab:13}
\end{center}
\vspace{-1.5em}
\end{table}

\begin{table}
\begin{center}
\caption{Performances of different methods on ICDAR15.}
\begin{tabular}{|c|c|c|c|c|}
  \hline
  Method & Recall & Precision & F-measure \\
  \hline
  \hline

  RRPN-2 \cite{ma2017arbitrary} & 72.65 & 68.53 & 70.53 \\
  seglink \cite{shi2017detecting} &73.1 & 76.8 & 75.0 \\

  Wordsup \cite{Hu17WordSup} & 77.03 & 79.33 & 78.16 \\
 
  EAST \cite{zhou2017east} & 78.33 & 83.27 & 80.78 \\

  He et al. \cite{he2017direct} & 80 & 82& 81 \\
  PixelLink \cite{deng2018pixellink} &  82.0 &85.5 & \textbf{83.7} \\
  \hline 
  CENet (VGG 50k+icdar15 finetune) & 79.2 & 86.1 & 82.5\\
  \hline
\end{tabular}
\label{tab:15}
\end{center}
\vspace{-1.5em}
\end{table}

\begin{table}
\begin{center}
\caption{Results of different methods on MSRA-TD500.}
\begin{tabular}{|c|c|c|c|c|}
  \hline
  Method & Recall & Precision & F-measure \\
  \hline
  \hline
Zhang et al.\cite{zhang2016multi} &67 &83 & 74 \\
EAST~\cite{zhou2017east} & 67.43 & 87.28 & 76.08 \\
He et al.\cite{he2017direct} &70 & 77 & 74 \\
PixelLink \cite{deng2018pixellink} &  83.0  &73.2 & 77.8 \\
  \hline
  CENet(VGG 50k+MSRA TD500 finetune) & 75.26 & 85.88 & \textbf{80.21} \\
  \hline
\end{tabular}
\label{tab:td500}
\end{center}
\vspace{-1.5em}
\end{table}

\begin{table}
\begin{center}
\caption{Performances on Total-Text dataset.}
\begin{tabular}{|c|c|c|c|c|}
  \hline
  Method & Recall & Precision & F-measure \\
  \hline
  \hline
  DeconvNet \cite{chng17tt} & 33 & 40 & 36 \\
  \hline
  CENet (VGG 50k) & 42.39 & 58.17 & 49.04\\
  CENet (VGG 50k+Total Text finetune) & 54.41 & 59.89 & \textbf{57.02}\\
  \hline
\end{tabular}
\label{tab:tt}
\end{center}
\vspace{-1.5em}
\end{table}

On the most challenging Total-text dataset, the proposed method presents an overwhelming advantage over other methods in comparison, as is shown in Tab 4. The baseline comes from DeconveNet that predicts a score map of text followed by connected component analysis. VGG 50K dataset contains some curved text, therefore the model trained on VGG 50k solely works reasonably well. With finetuning on the provided training set, our final model significantly outperforms the baseline method. Besides, this result strongly exhibits the effectiveness of our weak supervision mechanism for training the character detector. $s, d$ are set to 0.4 and 0.38 in this dataset.

We visualize detection results of our model on four benchmarks, illustrates in Fig. \ref{fig:res}. Results show our model can tackle text detection in various scenarios, especially on curved texts. 

\subsection{Future Works}


Our model predicts rich information including text level boundaries as well as character bounding boxes. With a view to these advantages, we hope to incorporate the acquired detection information into the follow-up text recognition. For instance, we may use the predicted character position to align the attention weight or boost CTC based recognition.




\section{Conclusion}
Observing the demerits of previous text detection methods, we present a novel scene text detection model. 
The model is more flexible to detect texts that captured unconstrained, 
the curved or severely distorted texts in particular. it is completely data-driven in an end-to-end way and thus makes little use of heuristic rules or handcrafted features. it is also trained with two correlated tasks, i.e., the character detection and character embedding, which is unprecedented. To train the network smoothly, we also propose several measures, i.e. weak supervision mechanism for training character detector and positive-negative pair reweighting, to facilitate training and boost the performance. 
Extensive experiments on benchmarks show that the proposed framework could achieve superior performances even though texts are displayed in multi-orientated, line-level or curved ways.
%
%
%
\bibliographystyle{splncs04}
\bibliography{egbib}

\end{document}